\documentclass[conference]{IEEEtran}

\usepackage{algorithmic}
\usepackage{amsmath,amssymb,amsfonts}
\usepackage{balance}
\usepackage{bm}
\usepackage{booktabs}
\usepackage{cite}
\usepackage{colortbl}
\usepackage{float}
\usepackage[T1]{fontenc}
\usepackage{graphicx}
\usepackage{subfigure}
\usepackage{mathtools}
\usepackage{multirow}
\usepackage{siunitx}
\usepackage{subfigure}
\usepackage{textcomp}
\usepackage{threeparttable}
\usepackage[color=green!40]{todonotes}
\usepackage{xcolor}

\usepackage{hyperref}
\graphicspath{{./}{figures/}}

\begin{document}

\title{%
  XJTLUIndoorLoc: A New Fingerprinting Database for Indoor Localization and
  Trajectory Estimation Based on Wi-Fi RSS and Geomagnetic Field%
}

\author{%
  \IEEEauthorblockN{Zhenghang Zhong, Zhe Tang, Xiangxing Li, Tiancheng Yuan,
    Yang Yang, Meng Wei, Yuanyuan Zhang, Renzhi Sheng,\\%
    Naomi Grant, Chongfeng Ling, Xintao Huan, Kyeong Soo Kim and Sanghyuk Lee}%
  \IEEEauthorblockA{%
    Department of Electrical and Electronic Engineering, Xi'an
    Jiaotong-Liverpool University, Suzhou, 215123, P. R. China.\\%
    (Correspondence: Kyeongsoo.Kim@xjtlu.edu.cn)%
  }%
}%

\maketitle

\begin{abstract}
  In this paper, we present a new location fingerprinting database comprised of
  Wi-Fi received signal strength (RSS) and geomagnetic field intensity measured
  with multiple devices at a multi-floor building in Xi'an Jiatong-Liverpool
  University, Suzhou, China. We also provide preliminary results of localization
  and trajectory estimation based on convolutional neural network (CNN) and long
  short-term memory (LSTM) network with this database. For localization, we map
  RSS data for a reference point to an image-like, two-dimensional array and
  then apply CNN which is popular in image and video analysis and
  recognition. For trajectory estimation, we use a modified random way point
  model to efficiently generate continuous step traces imitating human walking
  and train a stacked two-layer LSTM network with the generated data to remember
  the changing pattern of geomagnetic field intensity against $(x, y)$
  coordinates. Experimental results demonstrate the usefulness of our new
  database and the feasibility of the CNN and LSTM-based localization and
  trajectory estimation with the database.
\end{abstract}
\begin{IEEEkeywords}
  Indoor localization, trajectory estimation, received signal strength, Wi-Fi fingerprinting, deep
  learning, CNN, LSTM,  geomagnetic field.
\end{IEEEkeywords}
\section{Introduction}
\label{sec:xjtlu-camp-inform}
With the increasing demands for location-aware services and proliferation of
smart phones with embedded high-precision sensors, indoor localization has
attracted lots of attention from the research community. Global navigation
satellite system (GNSS) like global positioning system (GPS), which provides
accurate geo-spatial positioning, cannot be used indoors as the radio signals
from satellites is easily blocked in an indoor environment. Because Wi-Fi signal
and geomagnetic filed are widely available in an indoor environment, on the
other hand, indoor localization based on them through location fingerprinting
technique becomes popular. However, there are few available databases which
combines both Wi-Fi signal and geomagnetic field, which is why majority of
indoor localization schemes are based on either type of the data but not
both. XJTLUIndoorLoc, the new location fingerprinting database we present in
this paper, is another open database comprised of both Wi-Fi received signal
strength (RSS) and geomagnetic field intensity and our attempt to encourage more
research on indoor localization based on location fingerprinting with diverse
types of data.

Note that there are a few publicly available fingerprinting databases. Among
them, the RSS-based UJIIndoorLoc database \cite{torres-sospedra14:_ujiin} is
well known as the largest and the first publicly-available database. Two other
databases comparable to XJTLUIndoorLoc are UJIIndoorLoc-Mag
\cite{torres-sospedra15:_ujiin_mag} based on magnetic field and IPIN 2016
databases \cite{barsocchi16:_wlan} supporting Wi-Fi fingerprinting, geomagnetic
coordinates and inertial measurement units (IMU) data. As for sample size,
UJIIndoorLoc has 933 reference points, while UJIIndoorLoc-Mag and IPIN 2016
database have 281 and 325 reference points, respectively. In comparison,
XJTLUIndoorLoc has 969 reference points and provides RSS values, geomagnetic
coordinates, and IMUs like IPIN 2016 database, which put XJTLUIndoorLoc in a
unique position in terms of database size and the variety of fingerprinting
data.

Compared to UJIIndoorLoc-Mag and IPIN 2006 databases collecting accelerometer
coordinates and orientation coordinates, the data in XJTLUIndoorLoc were
collected with smartphones heading to four directions at each reference
point. The number of devices used for XJTLUIndoorLoc (i.e., two smartphones) is
comparable to UJIIndoorLoc-Mag (i.e., two smartphones) and IPIN 2006 (i.e., two
smartphones and one smartwatch). As for coverage, all databases are quite
different from one another. UJIIndoorLoc covers three buildings with 4 or 5
floors each, while UJIIndoorLoc-Mag and IPIN 2016 databases covers a single
laboratory room with 8 different corridors and a single floor of a building,
respectively. XJTLUIndoorLoc covers the 4th and 5th floors of the International
Business School Suzhou (IBSS) on XJTLU.

To demonstrate the use of the XJTLUIndoorLoc database, we also provide
preliminary results of localization and trajectory estimation based on
convolutional neural network (CNN) and long short-term memory (LSTM)
network. For localization, we consider mapping of unstructured RSS data for a
given reference point to an image-like, two-dimensional array and then apply CNN
tailored for image processing and recognition. For trajectory estimation, we use
a modified random way point (RWP) model to efficiently generate continuous step
traces imitating human walking and train a stacked two-layer LSTM network with
the generated data to remember the changing pattern of geomagnetic field
intensity against $(x, y)$ coordinates.


\section{XJTLUIndoorLoc: A New Fingerprint Database}
\label{sec:xjtl-new-fing}

\subsection{Data collection}
\begin{figure}[!htb]
  \setlength{\abovecaptionskip}{0.cm}%
  \setlength{\belowcaptionskip}{-0.cm}%
  \begin{minipage}[t]{0.29\linewidth}
    \begin{center}
      \includegraphics[width=\linewidth]{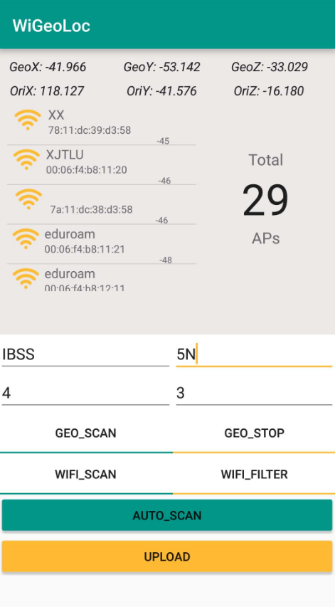}\\
      {\scriptsize (a)}
    \end{center}
  \end{minipage}%
  \hfill
  \begin{minipage}[t]{0.69\linewidth}
    \begin{center}
      \includegraphics[width=.8\linewidth]{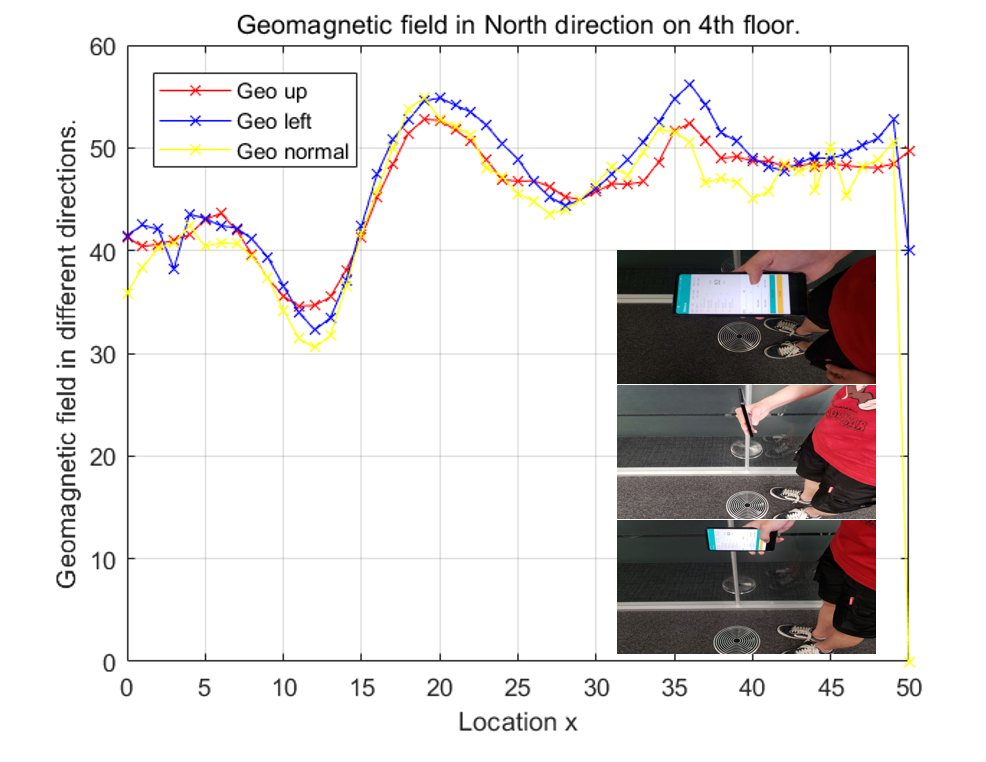}\\
      {\scriptsize (b)}
    \end{center}
  \end{minipage}
  \caption{Android App for data measurement: (a) Screen shot and (b) different
    postures of measuring data.}
  \label{fig:app}
\end{figure}

\begin{figure}[!htb]
  \setlength{\abovecaptionskip}{0.cm}
  \setlength{\belowcaptionskip}{-0.cm}
  \centering
  \subfigure[Y=1]{
    \includegraphics[width=0.2\textwidth]{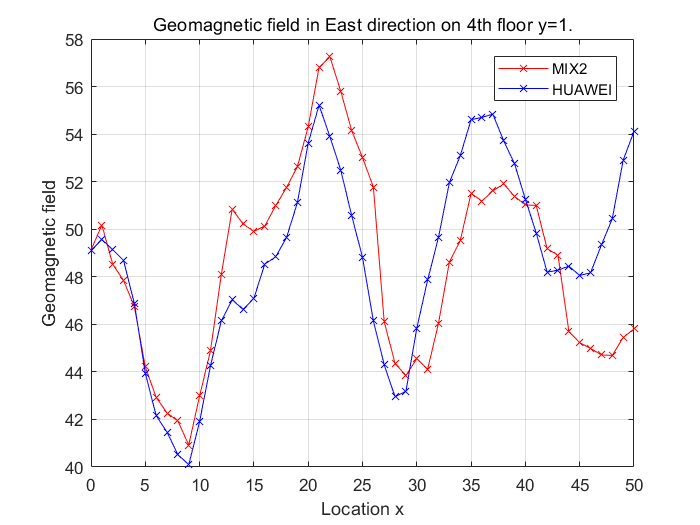}
  }
  \subfigure[Y=2]{
    \includegraphics[width=0.2\textwidth]{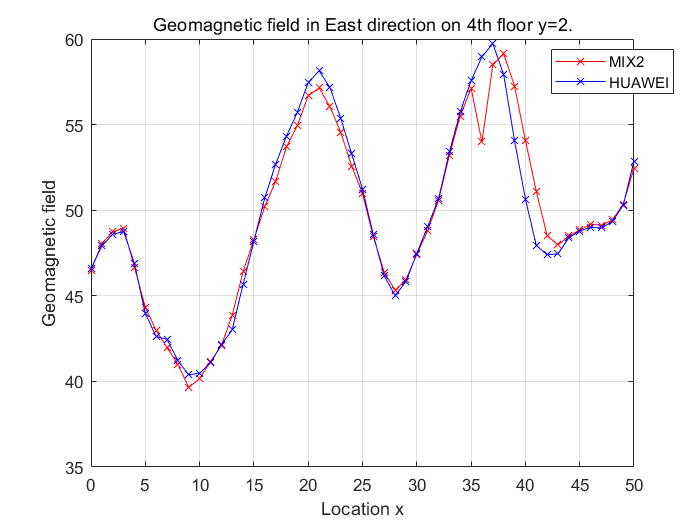}
  }
    \subfigure[Y=3]{
      \includegraphics[width=0.2\textwidth]{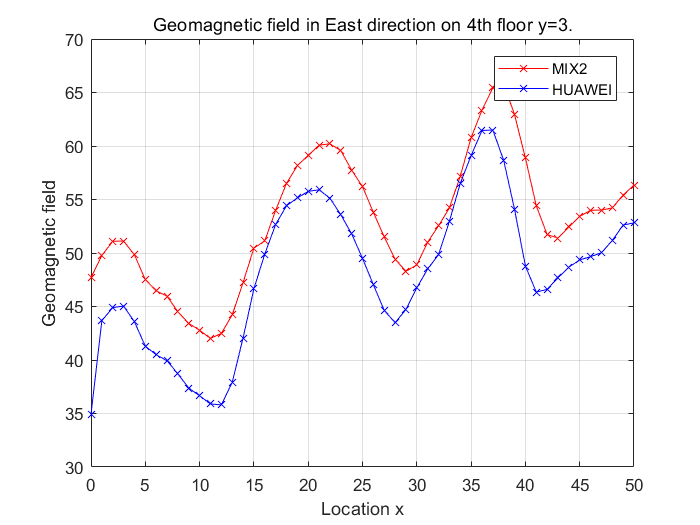}
    }
    \subfigure[Y=4]{
      \includegraphics[width=0.2\textwidth]{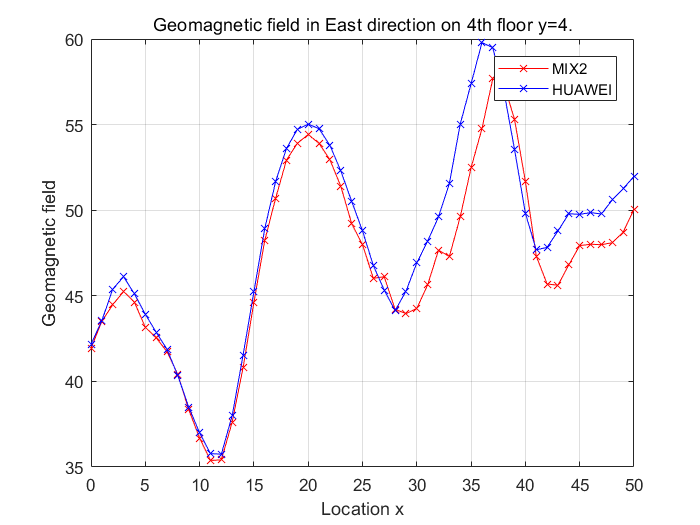}
    }
  \caption{Geomagnetic field in East direction on 4th floor y=[1, 4].}
  \label{fig:subfig} 
\end{figure}

We developed Android App named \textit{WiGeoLoc} to measure the fingerprint
data. Fig.~\ref{fig:app}~(a) shows its screen shot. The values of the
geomagnetic field and the angle of the orientation are on the top of the
screen. Before starting measurement, the location of the devices is requested to
input. Users can build the relative coordinate system to locate the
position. After the scanning, the numbers of the access point (AP) and the Wi-Fi
list will be shown on the screen. The application also supports data
transmission between server using the upload button. The data collected by the
application includes three parts, measuring of Wi-Fi and geomagnetic field part,
input location part and the device part, which has the time stamp and the brands
of the different measuring mobile phones. All data uses Android's official read
function. The different postures of the measuring mobile phones could also have
an impact on the outcome. In this work, left, horizontal and vertical
directions. Fig.~\ref{fig:app}~(b) shows different postures of measuring mobile
phone along the same measuring path and the changes in outcome. To testify the validness of collecting data by different sensors, Fig.~\ref{fig:subfig} displays geomagnetic field intensity collected by two sensors along four paths.

\subsubsection{Kalman Filter}
Considering the shaking of hands when detecting geomagnetic filed and
acceleration data, we implemented the Kalman Filter \cite{kalman60} in the
android part, to linearly filtering out the abnormal values.

\subsubsection{Mapping}
\begin{figure}[!htb]
\setlength{\abovecaptionskip}{0.cm}
\setlength{\belowcaptionskip}{-0.cm}
	\centering
	\subfigure[Plane graph]{
		\includegraphics[width=0.2\textwidth]{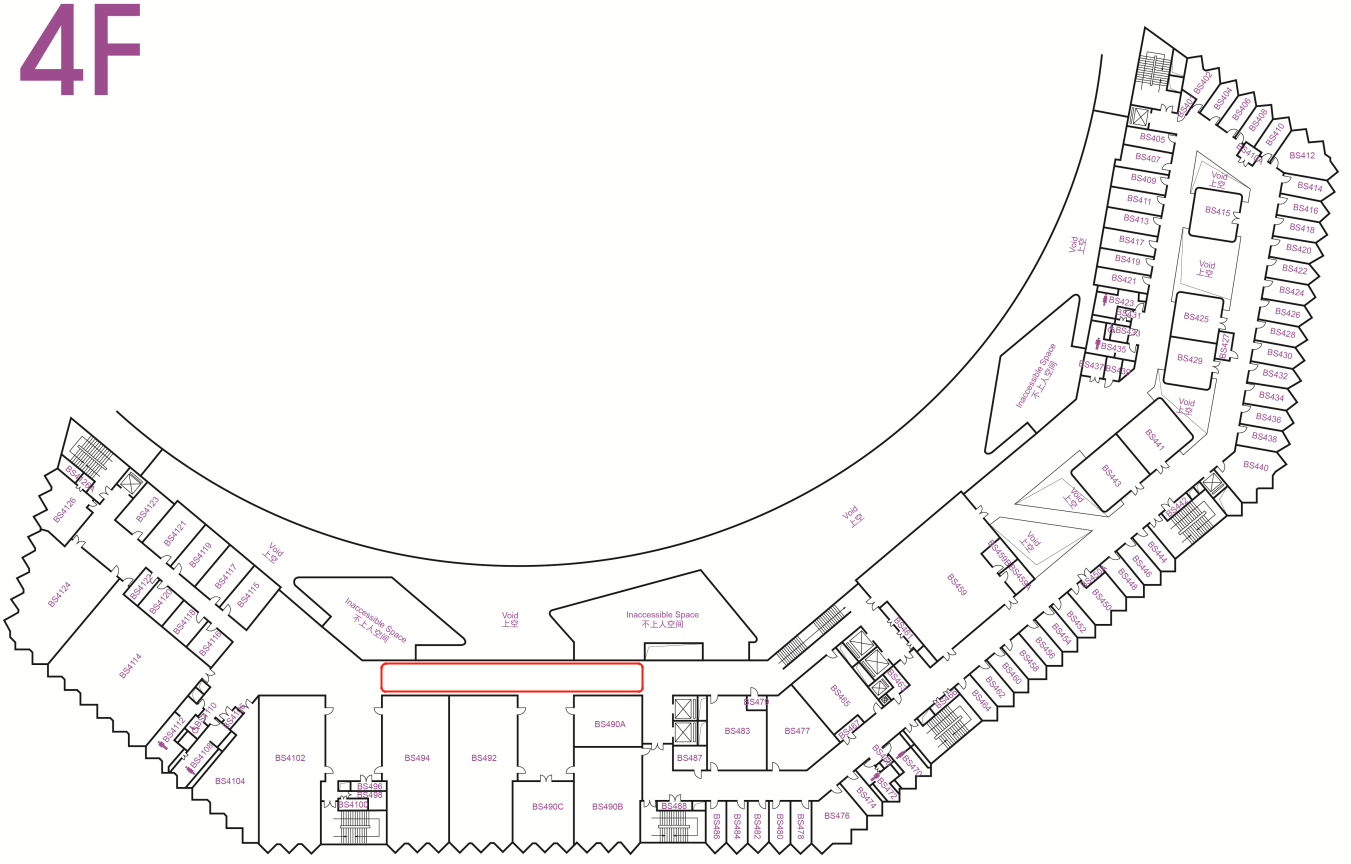}
	}
	\subfigure[Scene graph]{
		\includegraphics[width=0.18\textwidth]{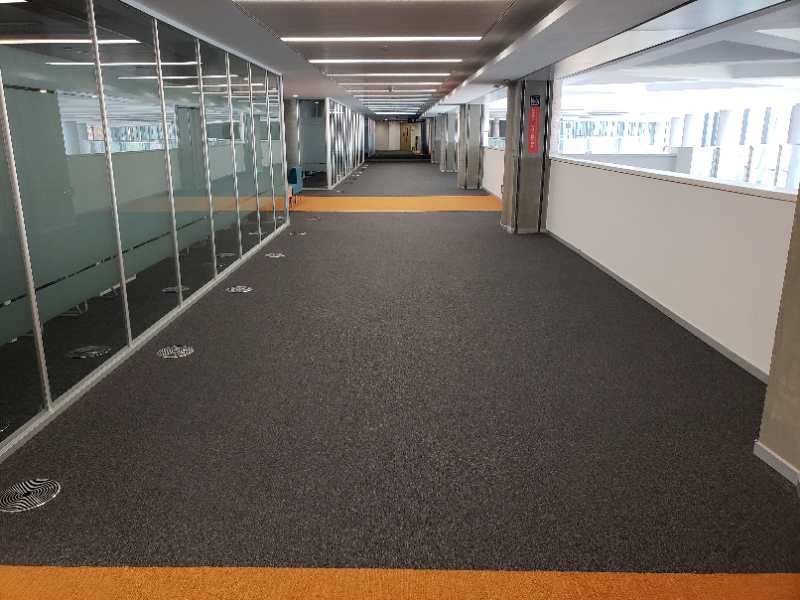}
	}
	\caption{IBSS 4th floor space structure.}
	\label{fig:IBSS_4} 
\end{figure}

In the fourth floor, the floor map can be seen in Fig.~\ref{fig:IBSS_4}. We
choose a rectangle area as the test bed with \SI{30}{\meter} in length and
\SI{7.2}{\meter} in width, which represent by red lines in Fig.~\ref{fig:IBSS_4}
(a). The rectangle scene is shown in Fig.~\ref{fig:IBSS_4} (b).

\begin{figure}[!htb]
\setlength{\abovecaptionskip}{0.cm}
\setlength{\belowcaptionskip}{-0.cm}
	\centering
    \vspace{-0.3cm}
	\subfigure[Plane graph]{
		\includegraphics[width=0.2\textwidth]{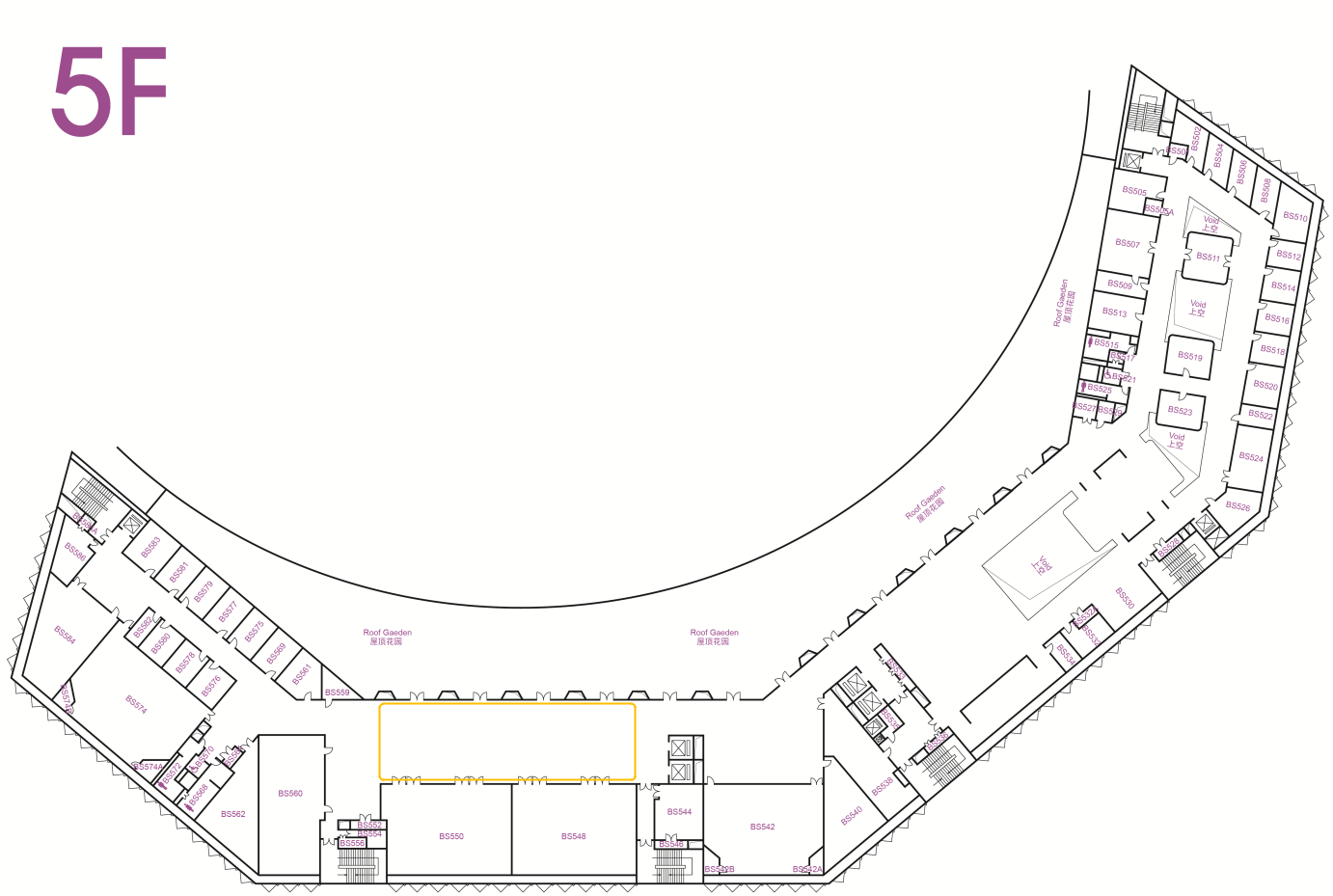}
	}
	\subfigure[Scene graph]{
		\includegraphics[width=0.2\textwidth]{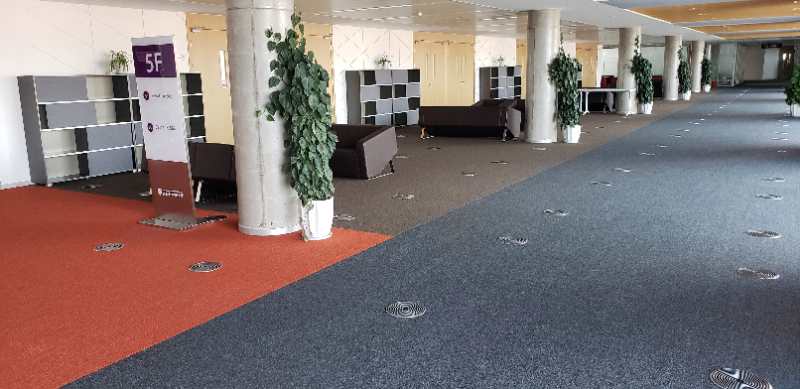}
	}
	\caption{IBSS 5th floor space structure.}
	\label{fig:IBSS_5} 
\end{figure}

In the fifth floor, the floor map is shown in Fig.~\ref{fig:IBSS_5}. The test
bed we chose is a rectangle area which measures \SI{30}{\meter} in length and
\SI{3}{\meter} in width, and the test bed is represented by blue lines in
Fig.~\ref{fig:IBSS_5} (a). The chosen scene is shown in Fig.~\ref{fig:IBSS_5}
(b).

\begin{figure}[!htb]
	\setlength{\abovecaptionskip}{0.cm}
	\setlength{\belowcaptionskip}{-0.cm}
  \centering
    \includegraphics[angle=0,width=0.6\linewidth]{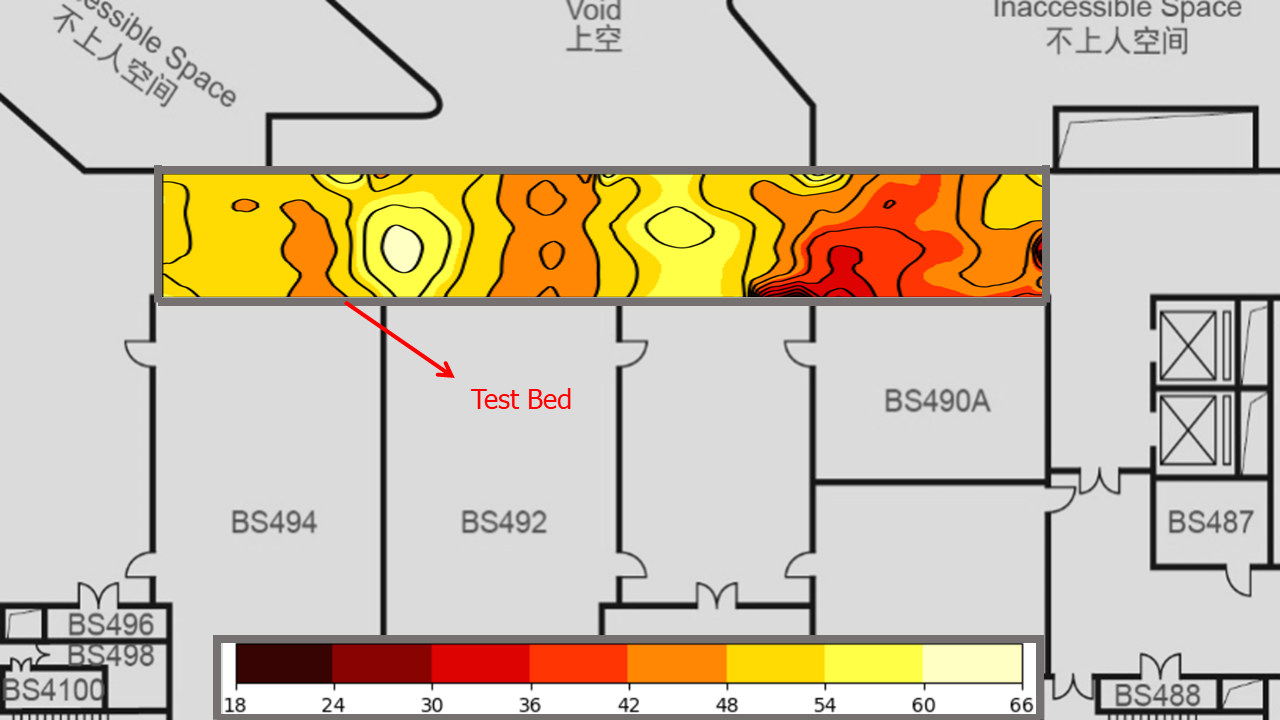}
    \includegraphics[angle=0,width=0.6\linewidth]{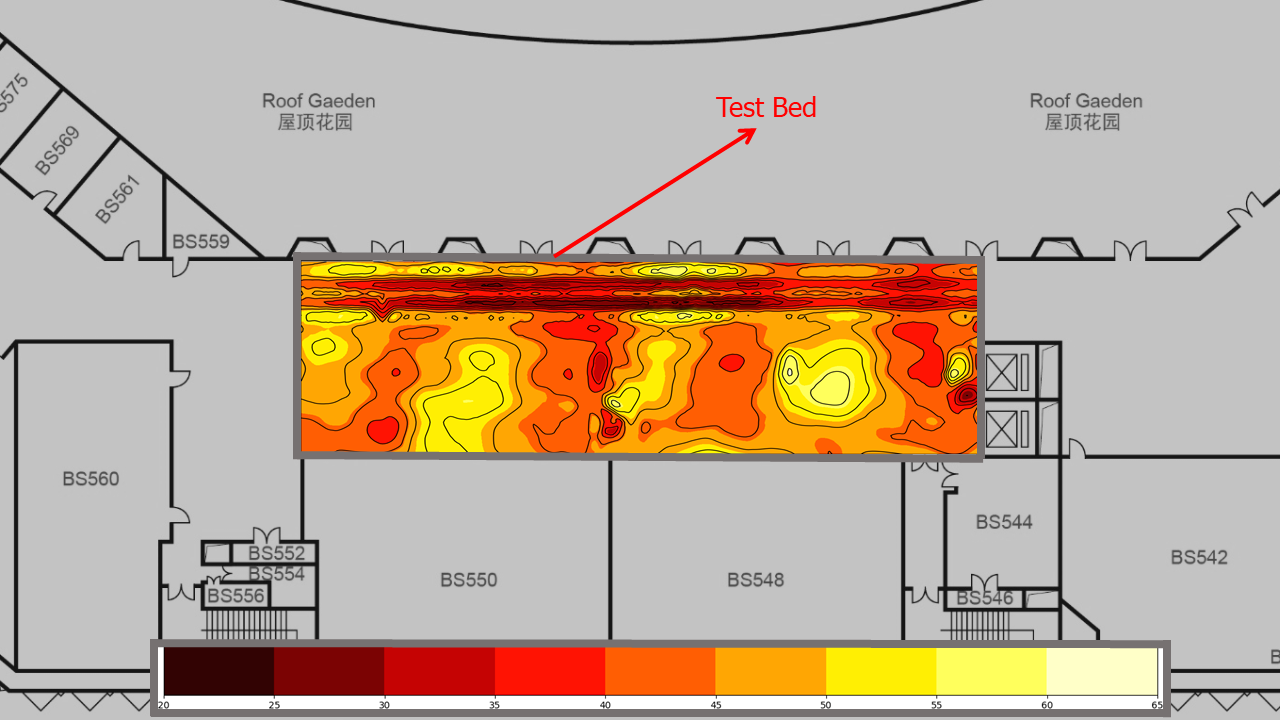}

    \caption{Geomagnetic field map of the fourth and the fifth floor in the IBSS
      Building at XJTLU.}
  \label{fig:floor_IBSS}
\end{figure}

In the two test beds, we build the $(x, y)$ coordinates and collect the
geomagnetic field data every \SI{60}{\cm}. Therefore, we create 306 reference
points on the fourth floor and 663 reference points on the fourth floor, which
means we totally create 969 reference points in the intersections of grid-layout
in the map. At each reference points, we measure four directions and on the
fifth floor, we use only one smart phone while on the fourth floor, we use two
smart phones. In addition, at the last line of the fourth floor, we add three
more directions, which are up, left and right. In order to improve the
resolution of the geomagnetic field map, we divide each \SI{60}{\cm} interval
into six 10-\si{\cm} interval and finally get 31,304 reference points using
Clough-Tocher Interpolation \cite{alfeld84} method in 2-dimensions, as it tends
to give a smooth interpolating surface and has a good interpolation speed. We
use two smart phones with the geomagnetic field measurement APP. The measured
data will upload to the server during the measurement. The final geomagnetic
field map of the fourth and the fifth floor are shown in
Fig.~\ref{fig:floor_IBSS}.

\subsubsection{Orientation}
\begin{figure}[!htb]
\setlength{\abovecaptionskip}{0.cm}
\setlength{\belowcaptionskip}{-0.cm}
  \centering
    \includegraphics[scale=.3]{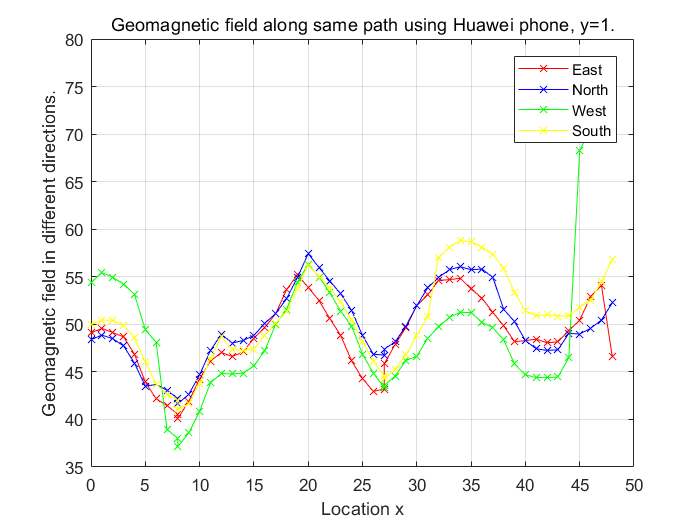}
  \caption{Geomagnetic field along same path in different attitude.}
  \label{fig:4Dir}
\end{figure}

In indoor magnetic environment, the strength of magnetic field which can be
detected by the magnetometer in Android mobile phone, is a three-dimensional
magnetic signal. Since the signal is relative to coordinate system of
magnetometer, and there existing intersection angle between reference system and
global coordinate system, the device would require to be fixed attitude to track
constant time variation. In the process of data collection, at each reference
point, the device was set horizontally and toward to north, west, east and south
direction, and Fig.~\ref{fig:4Dir} shows a sample group of measuring data along
same path toward different directions. Meanwhile, the posture of device as
left, right and up were both set for comparison, which is mentioned before shown
as Fig.~\ref{fig:app}~(b).

\noindent
Take the device diversity in consideration, we found at the same reference
point, the magnetic density of x, y and z components is different by different
devices. Analyze the data from two mobile phone, the deviation is valid,
however, the variance ratio of magnetic density of each dimensional component is
the same in generally. Concluded from the data analysis, in the following
experiment, fully utilizing the variance ratio of magnetic to match the
trajectory can reduce error to a great extent.

\subsubsection{Random way point model}
\begin{figure}[!htb]
 \centering
   \includegraphics[scale=.3]{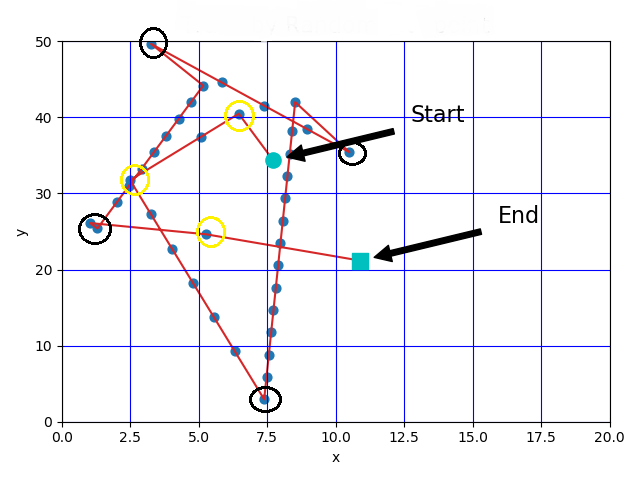}
 \caption{Trace generated based on RWP.}
 \label{fig:tra}
\end{figure}

RWP is the basic building block of most of the routing protocols and originally
used to simulate a communication protocol such as Ad Hoc networks
\cite{nayak15:_analy_manet_netsim}. However, in this work, students apply RWP to
generate 50,000 traces to model the movements of pedestrians, because RWP is a
random model for the movement of mobile nodes just like people's walking trace,
whose direction and speed are selected randomly and independently. This model is
a memory less mobility process where the information about previous status is
not considered for the future decision \cite{nayak15:_analy_manet_netsim}. To be
specific, our model has several input arguments that could be changed such as
the dimension of the map, the steps of each trace, the velocity range of each
step and the maximum pause time. In addition, the change of direction and
velocity only happens after every pause time. The output of RWP is a trace
represented by an array which contain the node positions in each step and the
illustration is shown in Fig. ~\ref{fig:tra}.
 
To train the LSTM by these traces, each point on the traces should have a
corresponding Geomagnetic filed strength. Because the traces are randomly
generated, students need to correspond all the points on the map to their own
Geomagnetic filed strength. However, it is impossible to get the strength of all
points by measurement. Thus, the map has been divided into many squares
(\SI[product-units = single]{60 x 60}{\cm}) and the Geomagnetic filed strength
will be measured on four terminal points. Then, students use Clough-Tocher
interpolation to simulate the Geomagnetic filed strength of any given
points. After the traces plotted, an improvement of RWP model has been found,
which can overcome the inherent defects of the model and making it possible to
improve the accuracy of the simulation that will be expounded in the Discussions
part.

\subsection{Data set description}
The database covers the lobby on 5th floor and the corridor on 4th floor. There
are totally 515 APs detected by Huawei P9. For each reference point, the
wireless accessed point, the location, the geomagnetic field strength and the
rotation angle are recorded as shown in table.\ref{tab:addlabel1} and
table.\ref{tab:addlabel2}. In addition, mobile device Huawei P9 and MiX2 were
detected toward north, south, east and west direction in both floor, but on 4th
floor, the attitude of device as left, right and up were both set for
measurement as previous mentioned.
\begin{table}[!htb]
\scriptsize
	\setlength{\abovecaptionskip}{0.cm}
	\setlength{\belowcaptionskip}{-0.cm}
    \vspace{0.05cm}
  \centering
  \caption{Database Structure.}
    \begin{tabular}{|c|c|c|c|c|c|c|}
    \toprule
    WAP000 & \textbf{...} & WAP515 & Loc\_x & Loc\_y & Floor & Building \\
    \midrule
    -110  & \textbf{...} & -110  & 0     & 0     & 5E    & IBSS \\
    \midrule
    -110  & \textbf{...} & -110  & 1     & 0     & 5E    & IBSS \\
    \midrule
    -110  & \textbf{...} & -110  & 2     & 0     & 5E    & IBSS \\
    \midrule
    -110  & \textbf{...} & -110  & 3     & 0     & 5E    & IBSS \\
    \midrule
    -110  & \textbf{...} & -110  & 4     & 0     & 5E    & IBSS \\
    \midrule
    -110  & \textbf{...} & -110  & 5     & 0     & 5E    & IBSS \\
    \bottomrule
    \end{tabular}%
  \label{tab:addlabel1}%
\end{table}%
\begin{table}[!htb]
\scriptsize2
	\setlength{\abovecaptionskip}{0.cm}
	\setlength{\belowcaptionskip}{-0.cm}
  \centering
  \caption{Database continue from table I.}
    \begin{tabular}{|c|c|c|c|c|c|}
    \toprule
    GeoX  & GeoY  & GeoZ  & OriX  & OriY  & OriZ \\
    \midrule
    -25.6125 & -5.79286 & -29.9464 & 97.59351 & -4.38194 & -2.16679 \\
    \midrule
    -25.2571 & -5.475 & -29.8786 & 97.82641 & -3.71709 & -1.29526 \\
    \midrule
    -22.099 & -4.42014 & -29.9931 & 100.7725 & -0.04106 & -1.98857 \\
    \midrule
    -23.4641 & -5.41875 & -28.0094 & 102.3195 & -0.39816 & -0.95255 \\
    \midrule
    -23.8958 & -4.8006 & -26.4107 & 101.139 & -0.04733 & -0.86068 \\
    \midrule
    -24.7422 & -5.01172 & -25.4219 & 101.2211 & 0.013164 & -1.21606 \\
    \bottomrule
    \end{tabular}%
  \label{tab:addlabel2}%
\end{table}%
\section{CNN-Based Indoor Localization with RSS}
\label{sec:cnn-based-indoor}
CNN is a class of feed-forward artificial neural network, which can identify
two-dimensional graphics of displacement, scaling and other forms of distortion
invariance. CNN has been applied to on image and video processing, natural
language processing \cite{collobert08}, and recommender systems
\cite{oord13:_deep}.
Because CNN was originally used for image recognition, we consider the mapping
of unstructured RSS data into an image-like, two-dimensional array by cleverly
arranging APs to improve localization performance. \cite{ibrahim18:_cnn_rss}
proposed a method to arrange RSS into input matrix based on time
series. Although the method improves the accuracy, the size of RSS image is too
small to train deep neural network (DNN) model. Moreover, the method to collect
data is not convenient. To address the problem of input data set, therefore, we
propose a new CNN model with large input data set, and strong correlation between
each AP in input matrix.

\subsection{Algorithm and structure}
\begin{figure}[!htb]
	\setlength{\abovecaptionskip}{0.cm}
	\setlength{\belowcaptionskip}{-0cm}
	\centering
		\includegraphics[scale=0.5]{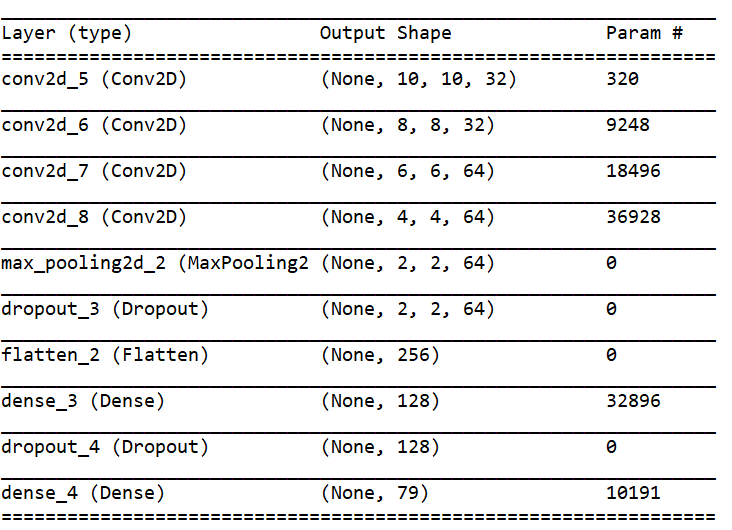}
	\caption{Convolution Neural Network Structure.}
	\label{fig:CNN_Structure}
\end{figure}

In the proposed algorithm, CNN is trained for localization using our RSS
fingerprint database. Comparing to other types of contemporary neural network,
CNN has overall better performance when image input is applied. The network
receives the pixels of the image as the input data. Therefore, CNN can learn the
features of image data better than traditional neural network. Additionally, CNN
can learn with data augmentation, which means the overall result of CNN will
remain unchanged if the original input image is shifted horizontally or
vertically, zoomed proportionally or flipped. A common CNN should contain the
following layers: convolution layers, activation layers, pooling layers and
fully connected layers and softmax. Fig.~\ref{fig:CNN_Structure} shows this
structure.

This CNN consists of 4 convolution layers, 1 max-pooling layer before it reaches
the fully connected layers.  The following 2 dropouts were applied to improve
the overall accuracy of CNN results which is also one of the proposed
optimization methods. After the image-like 2D matrix is accepted by CNN, the
first layer to operates data is convolution layer.  In this layer, another
smaller 2D matrix is exploited as the convolution kernel.  The functionality of
the kernels varies based on the value of weights in the matrix, for instance,
averaging, edge detection, etc.  The pooling layer is placed next to the
activation of convolution layer.  This layer can extract features of the
convolution layer output by reducing the number of rows and columns of the
matrix.  In this work, max pooling layer with a two by two filter (stride two)
will store the maximum value of the two by two subsection. At the final stage of
CNN, there are fully connected layers with softmax function which calculates the
output of CNN.  The softmax acts as a classifier based on the labels.

\subsection{Input data pre-processing}
\begin{figure}[!htb]
  \setlength{\abovecaptionskip}{0.cm}%
  \setlength{\belowcaptionskip}{-0.cm}%
  \centering
  \includegraphics[scale=1,angle=90]{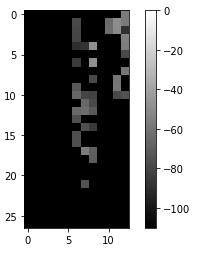}
  \caption{2D Input Image.}
  \label{fig:input1}
\end{figure}

The used CNN structure only accept 2D matrix as its image input.  Therefore, the
pre-processing of data is to convert the RSS data restored in this database into
a 2D matrix in a certain order.  The program should first acquire the proposed
database (in CSV format).  Then, rearrange these APs so the AP with largest RSS
value can be set to locate precisely in this reference point.  By repeating this
method for all APs in this database, all their relative locations can be fix on
a 2D image.  Consequently, a 2D feature image whose pixels are RSS value can be
created for each RP. One example of the image input is shown in
Fig.~\ref{fig:input1}.
The pixel on this grey scale image indicates the RSS value of each AP.

\section{LSTM-Based Indoor Localization with Geomagnetic Field}
\label{sec:lstm-based-indoor}
With the increase of Wi-Fi coverage area and signal intensity, Wi-Fi signal has
been a basis for indoor positioning. Since the large differences in Wi-Fi
signals at each point, Wi-Fi signals can be mapped to fingerprints. However, the
accuracy of localization based on Wi-Fi fingerprinting is not enough to provide
high quality location services \cite{jekabsons11:_wi_fi}. Compared to Wi-Fi
signal, geomagnetic field signal is time-invariant and less affected by RF
factor.
In addition, recurrent neural networks (RNNs) have been widely applied in the
area of indoor localization and trajectory estimation. The DeepML system
\cite{wang18:_deepm} combines magnetism and visible light for indoor
localization based on LSTM, and to fully take advantage of stability of visible
light. In \cite{jang17:_geomag}, authors train the RNN with traces generated
based on geomagnetic map, where the 3-dimensional magnetic field vector and
location coordinates compose the training data, and this innovation inspired us
with this experiment.

\subsection{Algorithm and structure}
In this work, the geomagnetic signal will be used for indoor localization. The
points in one trace are time dependent hence, using RNN like neural networks to
solve these problems would be an appropriate choice. The traditional RNN model
can focus on the previous information but for a long-term dependency problem
LSTM could be a solvent. For LSTM, each unit can remember the function of the
input and it can be kept for a long term, comparing with the RNN, the existence
of shortcut paths allowed the errors feedback, which can reduce the vanishing
gradients problem\cite{chung14:_empir_evaluat_gated_recur_neural}.

\begin{figure}[!htb]
  \begin{center}
  \includegraphics[angle=0,width=0.6\linewidth]{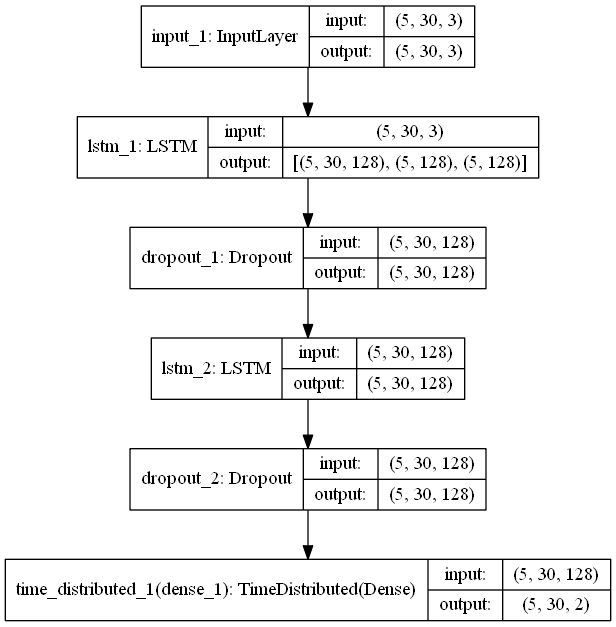}
  \end{center}
  \caption{Model structure of LSTM.}
  \label{fig:model_LSTM}
\end{figure}

We incorporate a deep LSTM to train the traces data, which is a state-of-art
recurrent neural network (RNN) to deal with long time dependencies
\cite{greff17:_lstm}.  We design a simple stacked stateful structure based on
LSTM, as shown in Fig.~\ref{fig:model_LSTM}, after the input layer, two stacked
LSTM layers are connected, and to prevent the overfitting we concatenate one
dropout layer after every LSTM layer. For the final layer we concatenate one
time-distributed dense layer, because we only want to interacting the values
between its own time step.

\subsection{Neural Network Training}
Before training we process the data into special shape, and based on a modified
RWP model, the generated trace of 100,000 steps is combined with its
3-dimensional geomagnetic field values based on Clough-Tocher 2-D
Interpolation. Considering the time step property in LSTM network, the sliding
window method is used to construct new shape of 3 dimensions, which are samples
times time steps times features, both for input and output data.

In the training process of RNNs, the number of batch size, time steps, hidden
nodes, the learning rate, and the total number of execution epochs are
significant parameters affecting the training performance.We vary all of these
parameters to find the parameter group of best performance. The input and output
data are normalized in terms of features. We use ADAM \cite{kingma17:_adam}
optimizer and mean square error (MSE) loss function. Then through the repeated
iteration of forward propagation and backpropagation based on the forward step
results, we optimize the relevant parameters.

\begin{figure}
  \setlength{\abovecaptionskip}{0.cm} \setlength{\belowcaptionskip}{-0.cm}
  \begin{minipage}[t]{0.5\linewidth}
    \centering
    \includegraphics[width=1.5in]{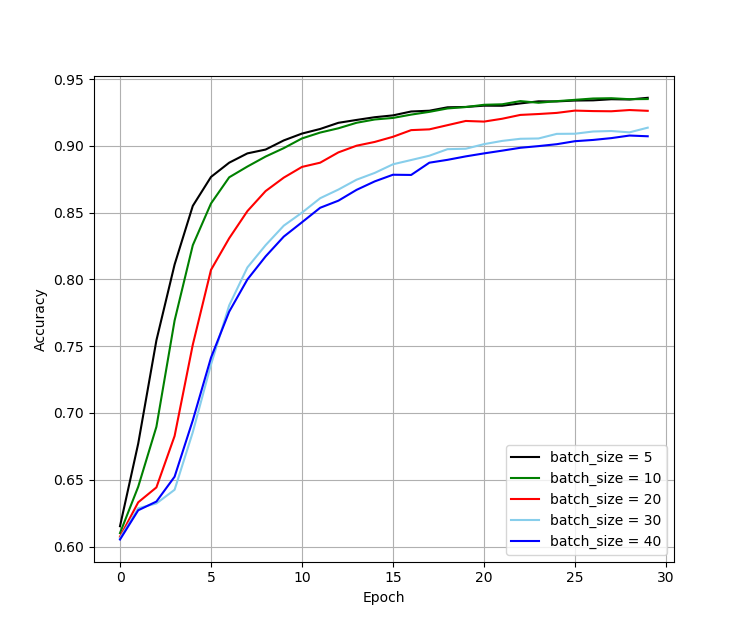}
    \caption{Localization accuracy of \protect\\training data in terms of
      number\protect\\ of batch size.}
    \label{fig:LSTM_diff_batch_size}
  \end{minipage}%
  \begin{minipage}[t]{0.5\linewidth}
    \centering
    \includegraphics[width=1.6in]{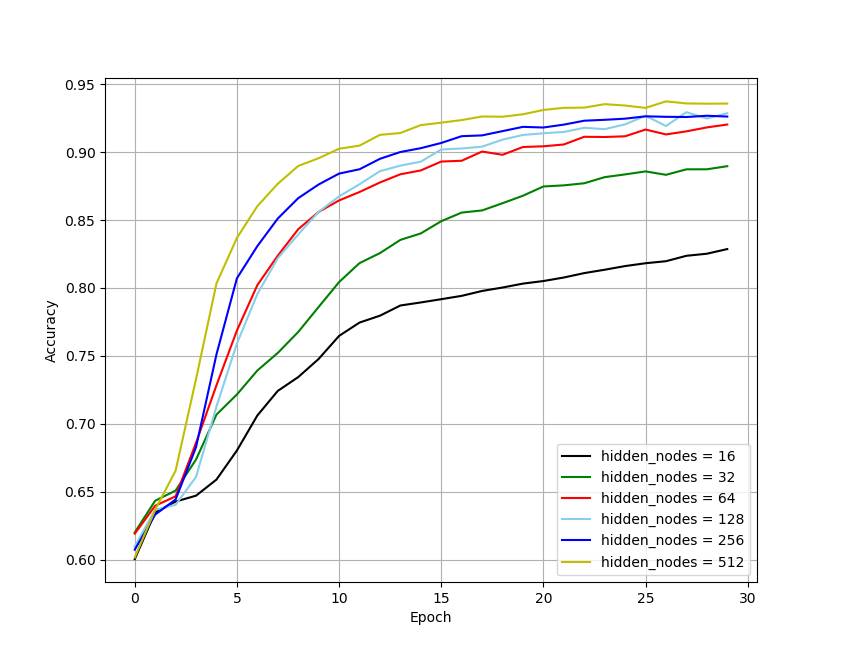}
    \caption{Localization accuracy of training data in terms of number of hidden
      node.}
    \label{fig:LSTM_diff_hidden_nodes}
  \end{minipage}
\end{figure}

Fig.~\ref{fig:LSTM_diff_batch_size} shows the localization accuracy of different
batch sizes. Rather than configure the input data as a single batch, which has a
worse performance in remembering previous data and gradient explosion can occur
for a large batch size. Fig.~\ref{fig:LSTM_diff_hidden_nodes} shows the
localization accuracy during the training in terms of epochs with varying hidden
nodes from 16 to 32, 64, 128, 256 and 512 with batch size of 20. We found out
that as we increase the number of hidden nodes, the accuracy performance could
generally be improved. However, we found that overfitting happen when the hidden
nodes reach a big value, which will be explained next.

\begin{figure}[tb]
  \centering
  \includegraphics[angle=0,width=0.7\linewidth,trim=7 5 10 10,clip=true]{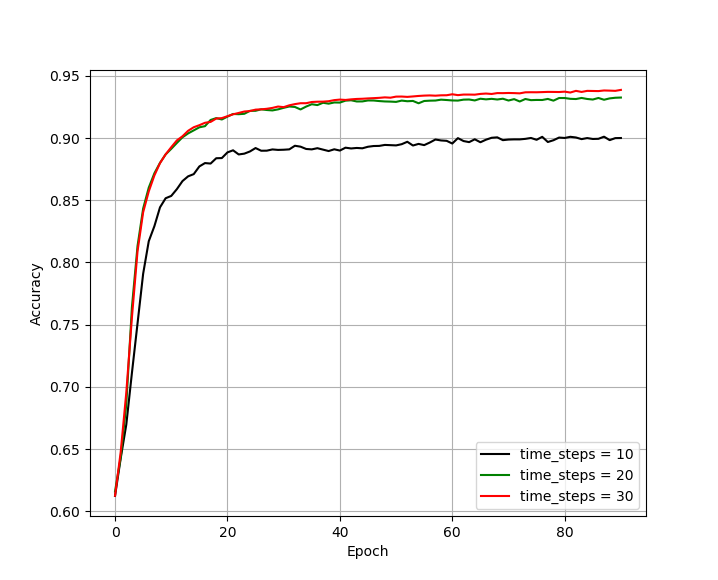}
  \caption{Localization accuracy of training data in terms of number of time
    steps.}
  \label{fig:LSTM_diff_steps}
\end{figure}

Fig.~.\ref{fig:LSTM_diff_steps} shows the localization accuracy during the
training in terms of different time steps, the overall performance increases
with increasing time steps, but tradeoff is made considering practical situation
and the increase in calculation burden. Accuracy of around 0.94 is reached with
time step of 30 while a small time step as 10 could only reach accuracy of
around 0.90.

Table~\ref{tbl:lstm_parameters} shows the configuration of the final LSTM model
optimized based on the training data and relevant parameters we set.
\begin{table}[!htb]
  \caption{LSTM Parameter Values.}
  \label{tbl:lstm_parameters}
  \centering
  \begin{tabular}{|l|l|}
    \hline
    \multicolumn{1}{|c|}{LSTM Parameter} & \multicolumn{1}{c|}{Value} \\ \hline\hline
    Ratio of Training Data to Overall Data & 0.75 \\
    Number of Epochs & 100 \\
    Batch Size & 5 \\ \hline
    Time Steps & 30 \\
    Hidden Nodes & 128 \\
    Optimizer & ADAM \cite{kingma17:_adam} \\
    Loss & Mean Squared Error (MSE) \\
    Dropout Rate & 0.2\\ \hline
  \end{tabular}
\end{table}

\section{Discussions}
\label{sec:discussions}

\subsection{XJTLUIndoorLoc Database}
\begin{figure}[!htb]
  \setlength{\abovecaptionskip}{0.cm} \setlength{\belowcaptionskip}{-0.cm}
  \centering \subfigure[4th floor]{
    \includegraphics[width=0.2\textwidth]{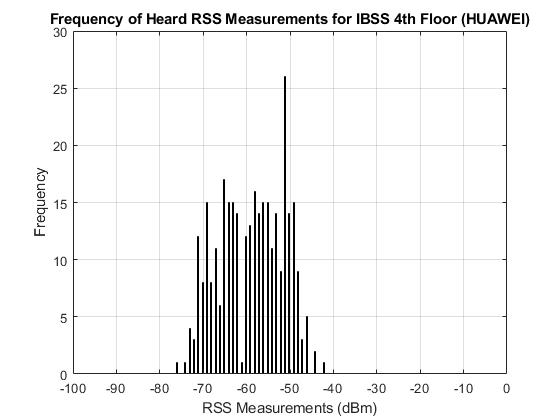}
  } \subfigure[5th floor]{
    \includegraphics[width=0.2\textwidth]{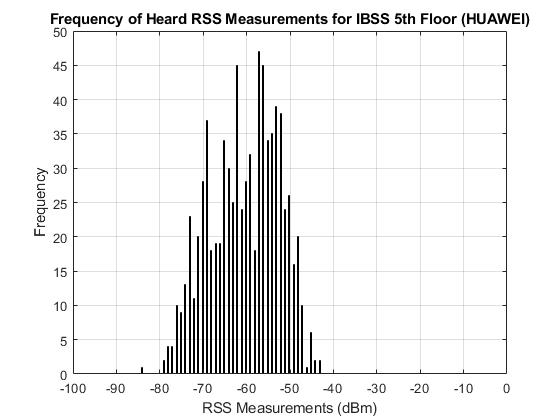}
  }
  \caption{Frequency of each RSS measurement by Huawei smartphones on the 4th
    and 5th floor of IBSS building.}
  \label{fig:Heard_RSS} 
\end{figure}

Fig.~\ref{fig:Heard_RSS} illustrates the frequency of each RSS measurement
tested by Huawei smartphones on the two floors at IBSS building. Also, other
databases have problems with APs being lost in either training or test data set
due to a large timed gap between the two measurements for data sets, which is
not the case for XJTLUIndoorLoc with as minimum of time as possible between the
training and testing data measurements. Taking into account all these factors,
XJTLUIndoorLoc enables accurate indoor localization despite the size of the
area, different path directions, type of smart phone and the numbers and
locations of buildings/floors.

However, there are also a few issues with the database:
\begin{itemize}
\item \textit{Difference in geomagnetic field values}: The values obtained by
  the two smart phones were quite different from each other, though these values
  should be similar to each other.
\item \textit{Deviation in manual measurement}: The location of the smart phones
  may not match the reference points perfectly due to the varying heights
  of the users.
\item  \textit{The trained data used is one directional data}: Theoretically, the
  rotational coordinates are best obtained by using rotation matrix or
  orientation.
\item \textit{Inconsistency in using mobile devices}: The MIX2 was only used on
  the 4th floor while the HUAWEI was used on both the 4th and 5th floors.
\item \textit{Inconsistency with RSS Measurements}: The numbers of RSS
  measurements with the two smart phones used on the 4th floor are different as
  shown in Fig.~\ref{fig:Heard_RSS}.
\end{itemize}

\subsection{CNN-Based Indoor Localization with RSS}
\begin{figure}[!htb]
	\begin{center}
		\includegraphics[scale=0.7,trim=7 3 10 10,clip=true]{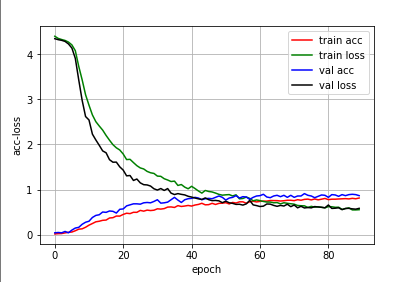}
	\end{center}
	\caption{The result curve of CNN experiment.}
	\label{fig:CNN_Result}
\end{figure}

The results of the CNN-based indoor localization was worse than the
state-of-the-art DNN-based indoor localization. Fig.~\ref{fig:CNN_Result} shows
the CNN results. There are possible explanations that should be discussed.  Our
database still need to be optimized because a major part of RSS stored in our
database is inconsistent.  The data fluctuates rapidly so the converted input
images were different even for the same reference point.  This prevents the
network from learning efficiently. Moreover, the computational complexity is
therefore raised to an extent where optimizing the algorithm will not grant
improvement on the accuracy.

If the problem of the database can be solved, there are some proposed method for
the optimization of CNN.  As is mentioned in methodology section, the addition
of dropout layer is effective.  Similarly, the batch normalization layer can be
added to reduce the computational complexity for max-pooling layer.  As a
result, the overall accuracy can be improved.  Furthermore, adjusting relevant
parameters (for instance weight decay, learning rate, etc.) can increase the
performance of CNN. Then, if the input image are translated, zoomed
proportionally or rotated and the network is trained with such data, the data
augmentation will lead to better results.

\subsection{LSTM-Based Indoor Localization with Geomagnetic Field}
In this work, the traces are created by the RWP. The speed and the direction
will influence the result of the RWP model. In order to mimic the human
trajectory indoor, the model needs to maximize the randomness. The result of the
traditional RWP model has the speed damping. An intuitive display is the
position of acute angle in trajectory. As shown in Fig.~\ref{fig:tra}, the
corners with black circles near the edge are sharper than the center corners,
which is the probability of large events. The reason is that the time cost by
the central points is more than edge points. This phenomenon causes the central
tendency of the trace. To solve the problem, an improvement of the RWP model,
which changed the node speed to satisfy the Gamma distribution, after changing
the speed decay can be overcome \cite{sabah10}. In Gamma RWP, a random uniform
distribution provides the initial position and direction of motion for each node
\cite{sabah10}. Due to time constraints, this model can be used in the future to
improve the personification of the trajectory, which may be helpful for training
accuracy.

\begin{figure}[H]
  \begin{center}
    \includegraphics[angle=0,width=0.6\linewidth]{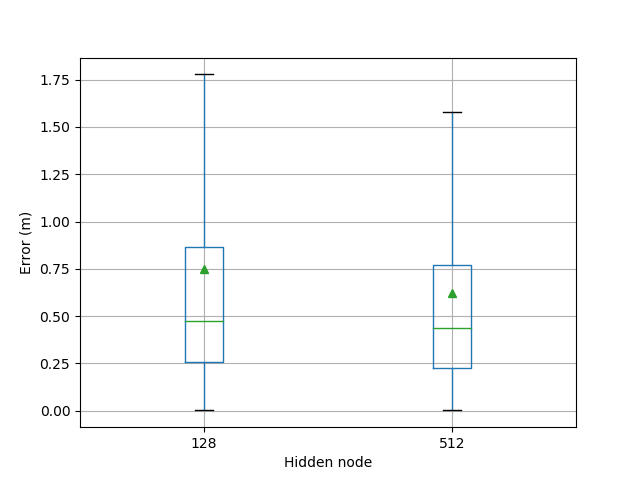}
  \end{center}
  \caption{Localization error in terms of hidden nodes.}
  \label{fig:error_LSTM}
\end{figure}

Fig.~\ref{fig:error_LSTM} shows the localization performance of the LSTM model
with varying hidden nodes from 128 to 512. The middle blue box for each element
in the figure represent the distribution of 75\% total dataset, while the top
and bottom line represent the distribution of 95\% of dataset. We can conclude
that with increasing hidden nodes, the general performance of the model would
increase, but due to the phenomenon of overfitting, a few points (remaining 5\%
of hidden dataset) would deviate a lot from actual value. Although the mean
error decrease from \SI{0.75}{\meter} to \SI{0.62}{\meter} when we increase the
number of hidden nodes from 128 to 512, we observe that the maximum error
increase from \SI{7.8}{\meter} to \SI{14.9}{\meter}. Moreover, for the
performance of 128 hidden nodes, mean of 75\% dataset is \SI{0.86}{\meter}. Due
to this overfitting problem, we choose the hidden nodes of 128 for a better
overall performance.

\section{Summary}
\label{sec:summary}
We have presented a new fingerprinting database called XJTLUIndoorLoc for indoor
localization and trajectory estimation based on Wi-Fi RSS and geomagnetic
field. XJTLUIndoorLoc, which covers a 216-\si{\meter\squared} lobby and a
90-\si{\meter\squared} corridor, contains 969 reference points. For geomagnetic
based localization, to improve the resolution of geomagnetic field map,
Clough-Tocher 2-D interpolation was applied to the samples for the lobby, which
increases the number of reference points from 663 to 21973. Then, we used RWP
model to model the traces which served as training data for LSTM. The training
result showed that the localization accuracy is highly aligned with the number
of batch size, the number of hidden nodes as well as the number of time
steps. Under optimum condition, the localization error over 25,000 samples reach
a mean value of \SI{0.7}{\meter}. We also described a new localization approach
based on CNN with image-mapped RSS data; the current results indicate that there
are lots of areas to improve. In the future, we plan to combine RSS and
geomagnetic field to extend our work in this paper to improve the performance of
indoor localization.

\section*{Acknowledgment}
This work was supported in part by Xi'an Jiaotong-Liverpool University (XJTLU)
Summer Undergraduate Research Fellowships programme (under Grant SURF-201830),
Research Development Fund (under Grant RDF-16-02-39), Research Institute for
Future Cities Research Leap Grant Programme 2016-2017 (under Grant RIFC2018-3),
and Centre for Smart Grid and Information Convergence. The floor maps of the
IBSS building were provided by XJTLU Campus Management Office.

\balance 


\end{document}